# Generative adversarial network based on chaotic time series

Makoto Naruse[1,2*], Takashi Matsubara[3], Nicolas Chauvet[1], Kazutaka Kanno[4], Tianyu Yang[2], and Atsushi Uchida[4]

[1] Department of Information Physics and Computing, Graduate School of Information Science and Technology, The University of Tokyo, 7-3-1 Hongo, Bunkyo-ku, Tokyo 113-8656, Japan

[2] Department of Mathematical Engineering and Information Physics, Faculty of Engineering, The University of Tokyo, 7-3-1 Hongo, Bunkyo-ku, Tokyo 113-8656, Japan

[3] Department of Computational Science, Graduate School of System Informatics, Kobe University, 1-1 Rokkodai, Nada, Kobe, Hyogo 657-8501, Japan

[4] Department of Information and Computer Sciences, Saitama University, 255 Shimo-Okubo, Sakura-ku, Saitama, Saitama 338-8570, Japan

\* Email: makoto_naruse@ipc.i.u-tokyo.ac.jp

**Abstract**

Generative adversarial network (GAN) is gaining increased importance in artificially constructing natural images and related functionalities wherein two networks called generator and discriminator are evolving through adversarial mechanisms. Using deep convolutional neural networks and related techniques, high-resolution, highly realistic scenes, human faces, among others have been generated. While GAN in general needs a large amount of genuine training data sets, it is noteworthy that vast amounts of pseudorandom numbers are required. Here we utilize chaotic time series generated experimentally by semiconductor lasers for the latent variables of GAN whereby the inherent nature of chaos can be reflected or transformed into the generated output data. We show that the similarity in proximity, which is a degree of robustness of the generated images with respects to a minute change in the input latent variables, is enhanced while the versatility as a whole is not severely degraded. Furthermore, we demonstrate that the surrogate chaos time series eliminates the signature of generated images that is originally observed corresponding to the negative autocorrelation inherent in the chaos sequence. We also discuss the impact of utilizing chaotic time series in retrieving images from the trained generator.

# Introduction

Generative adversarial network (GAN) is gaining increased importance in artificially constructing realistic images and related functionalities [1–7]. GAN is based on two networks called generator and discriminator, denoted by **G** and **D** respectively in Fig. 1a, which are evolving through adversarial learning [1]. Combined with deep convolutional neural networks (DCGAN) [2] and related techniques such as progressively growing scales (PGGAN) [3], amazingly realistic, high-resolution scenes and human faces have been successfully produced. The generator learns a map from latent space to data space, over which given samples are distributed, and the discriminator evaluates the map. Since the objective function is defined as the expected value of the discriminator output over the distribution of latent variables, the training requires vast amounts of pseudorandom numbers for Monte Carlo sampling of latent variables. In this study we examine GAN from a physics point of view; the impact of utilizing chaotic time series [8, 9] for GAN in the latent space.

Again, while GAN in general needs large and genuine training datasets, which is marked by **x** in Fig. 1a, it is noteworthy that vast amounts of pseudorandom numbers are required for the noise source, also called the latent variables, with respect to the generator as depicted by **Z** in Fig. 1a. In this study, we utilize chaotic time series which are experimentally generated by semiconductor lasers for the noise source [8–11] for GAN. Also, after finishing the learning phase, we examine the impact of utilizing chaotic time series subjected to the trained generator **G** as the latent variables in retrieving output images (Fig. 1b). We expect that the inherent nature of chaos, such as their time-domain correlations as well as irregularity, is transformed in the characteristics of the generated images. We show that the similarity in proximity, which is a degree of robustness with a minute change in the input latent variables, is enhanced while the versatility as a whole is not severely degraded. The impact of using chaotic sequences, instead of conventional pseudorandom numbers, in retrieving output images from the trained generator is also discussed. One additional remark in the beginning is that the technology of ultrafast optical random number generators (RNG) by chaotic lasers [10, 11] is not examined in this study; the primal interest of the present study is to highlight the impact of the utilization of experimentally observed chaotic sequences by lasers themselves [8, 9] for GAN applications.

## Result

The architecture of GAN utilized in this study is DCGAN [2] with the input image size of 64 × 64. The genuine data sets used for training are 202,559 kinds of human face images available from CelebA [8]. The network structures of the generator $G$ and the discriminator $D$ as well as the associated parameters are the same with the ones shown in Ref. [2]; for example, in the generator, a series of fractionally-strided convolutions followed by conversion to a 64 × 64-pixel image. A stochastic gradient descent method with a mini-batch size of 128 was used for training. The slope of the LeakyReLU was 0.2. Adam optimizer [13] with the learning rate of 0.0002 was used.

### Chaotic sequences

The input vector $\boldsymbol{Z}$ subjected to the generator $G$ consists of 100 elements of 32-bit floating point numbers between $-1$ and $1$ denoted by $\boldsymbol{Z} = (Z_1, \cdots, Z_i, \cdots, Z_{100})$ where $i = 1, \cdots, 100$. Unlike conventional uniformly-distributed pseudorandom numbers, we construct $\boldsymbol{Z}$ based on experimentally observed chaotic laser time series. A schematic illustration of generating chaotic time series is shown in the upper side of Fig. 1c where a fraction of the output light from the semiconductor laser is fed back to the cavity of the laser, accompanying certain time delay, leading to chaotic oscillation of the laser [8, 9]. Indeed, by exploiting the high-bandwidth attributes of light, ultrahigh-speed RNG have been demonstrated in the literature based on chaotic dynamics of semiconductor lasers [10, 11]. However, we should emphasize that, as mentioned in *Introduction*, while the adaptation of optical ultrafast RNG physically directly to deep learning signal processing hardware platform may be of great interests in future studies, the present study does *not* discuss optical RNG. In this study, we adapted the original laser chaos sequences, which were acquired separately prior to the computation process of GAN, to examine qualitative impacts of chaos in GAN.

Figure 1c represents examples of the chaotic signals where four kinds trains were shown referred to as (i) Chaos 1, (ii) Chaos 2, (iii) Chaos 3, and (iv) Chaos 4 acquired experimentally by slightly varying the reflection of the external mirror. The details are shown in the *Methods* section. Chaos 1, 2, 3, and 4 sequences were sampled by a high-speed digital oscilloscope at a rate of 100 GSample/s (10 ps sampling interval) with 10,000,000 (= 10M) points with an eight-bit resolution. Such 10M data points were stored 100 times for each signal train; hence, there were 100 kinds of 10M-long sequences;

we denote Chaos1$_i$(*t*), Chaos2$_i$(*t*), Chaos3$_i$(*t*), Chaos4$_i$(*t*) for the *t*-th sample of the total 10M points during the *i*-th measurement $(i = 1,\cdots,100)$.

We horizontally concatenate Chaos1$_i$(*t*) to Chaos4$_i$(*t*) to obtain a 32-bit integer variable by regarding each of 8-bit value from ChaosX$_i$(*t*) being an unsigned integer. It is then subtracted by $2^{32}$ followed by normalization so that the value ranges from −1 to 1. The resultant value is referred to as $Z_i$(*t*). In this manner, $Z_i$(*t*) is formed where *i* ranges from 1 to 100 and *t* spans from 1 to $10^7$.

In the present study, we examine seven kinds of sequences for the latent variable ***Z*** for GAN. The first is uniformly distributed pseudorandom numbers generated by Mersenne Twister, referred to as RAND hereafter. In Fig. 2(i, a), the first 100 points of time sequence $Z_1$(*i*) is displayed (*time*-domain snapshot) while Fig. 2(i, b) represents the array of $Z_i$ as a function of the index *i*; we call it a *space*-domain snapshot of ***Z***(*t*). The histogram of the signal level is shown in Fig. 2(i, c) where uniform distribution is confirmed. The column (d) and (e) of Fig. 2 demonstrates autocorrelation of ***Z***(*t*) in the temporal and spatial dimension, respectively. That is, Fig. 2d represents the temporal average of the autocorrelation of $Z_i$(*t*) with respect to *t* whereas Fig. 2e displays the sequence average of the autocorrelation of $Z_i$(*t*) with respect to *i* ranging from 1 to 100. Clearly, since there are no temporal or spatial correlations inherent in RAND, the autocorrelations are zero both in Figs. 2(i, d) and 2(i, e) except the case when the lag is zero.

With the chaotic sequence ***Z***(*t*), the temporal and spatial snapshots and the signal level histogram are shown respectively in Figs. 2(ii, a), 2(ii, b), and 2(ii, c) where the signal incidence exhibits peaky levels around −1 and 1. This is because, as discussed detail in later, the signal level distribution of the original chaotic lasers is similar to Gaussian distribution with zero being the average. In the above-mentioned construction of $Z(t)$, however, we deal with the 8-bit binary sequences as unsigned integer values; hence the Gaussian-like distribution split positive and negative values while the peaks appear at the edges (−1 and 1). The autocorrelation along the temporal dimension (Fig. 2(ii, d)) does exhibit non-zero values along the time lag; in particular, it shows negative maximum value of around −0.1 when the time lag is −5 or 5. Conversely, the autocorrelation along the space dimension (Fig. 2(ii, e)) is zero, manifesting that there is no correlation among different elements in $Z$. Since the sampling rate of the original chaotic lasers is 100 GSample/s, the signals in the row (ii) is referred to as "Chaos [Time-domain] (SI: 10 ps)" hereafter where SI stands for sampling interval.

As observed in the time-domain autocorrelation (Fig. 2(ii, d)), the time difference of 5 cycles, corresponding the sampling interval of 50 ps of the chaotic lasers, is a characteristic variable since it yields the negative maximum autocorrelation. The column (iii) of Fig. 2, denoted by "Chaos [Time-domain] (SI: 50 ps)", is the case when $Z(t)$ is resampled at every 50 ps. While the shape of the histogram (Fig. 2(iii, c)) is the same with the previous case (Fig. 2(ii, c)), the time-domain autocorrelation shows the negative maximums at the time lag of −1 and 1.

The data shown in the rows (iv) and (v) were normally distributed random variables with standard deviation being 0.1 and 0.2, referred to as "RANDN σ = 0.1" and "RANDN σ = 0.2" hereafter, respectively. In order to obtain resemblant distribution to Chaos (ii) and (iii), the positive and the negative parts are shifted toward the edge of −1 and 1, which is an equivalent transformation conducted in Chaos (ii) and (iii) above. The signal level distribution is similar to the case of chaos when σ = 0.2.

In addition, we prepare another arrangement of latent variables from chaos so that correlations are transformed into *space* domain of **Z**(*t*), instead of the former *time* domain arrangement represented by the assignment of chaotic sequences $Z_1(t)$, $Z_2(t)$, …, $Z_{100}(t)$ to **Z**(*t*) ($=[Z_1(t),\cdots,Z_{100}(t)]$). In the case of space domain assignment, the first and the second **Z**(*t*), for example, are given by $\mathbf{Z}(1)=[Z_1(1),\cdots,Z_1(100)]$ and $\mathbf{Z}(2)=[Z_1(101),\cdots,Z_1(200)]$, respectively, meaning that **Z** consists of the consecutive source of the same sequence (in this case $Z_1$). Therefore, correlation among the elements *within* a single $\mathbf{Z}(t)$ shows up. Generally, **Z**(*t*) is given by

$$\mathbf{Z}(t)=\left[Z_1\left(100\times(t-1)+1\right),\cdots,Z_1\left(100\times(t-1)+100\right)\right] \tag{1}$$

when *t* ranges from 1 to $10^5$; remember that $10^5$ is the number of data points of the original chaotic sequence. After $t=10^5+1$ to $t=2\times10^5+1$, the second sequence $Z_2(t)$ was utilized for **Z**(*t*);

$$\mathbf{Z}(t)=\left[Z_2\left(100\times(t-10^5-1)+1\right),\cdots,Z_2\left(100\times(t-10^5-1)+100\right)\right]. \tag{2}$$

In such manners, $Z_3(t)$, …, $Z_{100}(t)$ were subsequently allocated to **Z**(*t*). While observing the equivalent histogram to the former time-domain cases, the autocorrelation exhibits contrasting properties; the space-domain autocorrelation (Fig. 2(vi, e)) emerges while the time-domain one diminishes (Fig. 2(vi, d)). The row (vii) displays another the space-domain arrangement by setting the sampling interval being 50 ps, referred to as "Chaos [Space-domain] (SI: 50 ps)".

In the training phase, the number of iterations for an epoch is 2000. We examine the output of the resultant model of the generator after 20 epochs. The hardware used for this study is a personal computer environment (HPC Systems, CPU: Intel Xeon 3.0 GHz, RAM: 384 GB, Windows 10) with a single graphical processing unit (ELSA GeForce GTX 1050 Ti 4GB SP).

## Discussion

### Similarity in proximity

Figures 3a and 3b show representative examples of the pictures generated by (i) RAND and (iii) Chaos [Time-domain] SI: 50 ps, respectively, where we can observe equivalently natural human faces. There are certainly artifacts and unrealistic portions contained in the images; the latest sophisticated GAN techniques, such as the ones in Refs. [6, 7], may resolve such problems in the quality of the generated pictures. However, we proceeded the generation and the analysis based on the above-described DCGAN technique and utilizing our not-powerful computing environment since the primal interest of the present study is to examine the impact of chaotic sequences.

First, we examine the robustness of the resultant images with respect to the slight change of the input latent variables. With a randomly chosen reference points for the latent variables $\boldsymbol{Z}^{\text{REF}(k)} = (Z_1^{\text{REF}(k)}, \cdots, Z_{100}^{\text{REF}(k)})$ with $Z_i^{\text{REF}(k)}$ ($i = 1, \cdots, 100$) being generated by pseudorandom numbers, the reference output picture $\boldsymbol{P}^{(k)}$ is generated. Total $K = 200$ reference points of $\boldsymbol{Z}^{\text{REF}(k)}$ were chosen, of which ten cases are displayed in the left-hand side edge column framed by squares (Figs. 3c and 3d). With respect to a specific $\boldsymbol{Z}^{\text{REF}(k)}$, its proximity or neighbours are specified by $\mathbf{Z}^{N(k,l)} = \left( Z_1^{REF(k)}, \cdots, Z_{90}^{REF(k)}, Z_{91}^{N(l)}, \cdots, Z_{100}^{N(l)} \right)$ in which the first ninety elements are the same with $\mathbf{Z}^{REF(k)}$ whereas the last ten elements were randomly given by pseudorandom numbers. Such neighbours are arranged for $L = 100$ points; ten of such neighbouring pictures, $\mathbf{P}^{N(k,l)}$, are shown in the horizontal directions in Figs. 3c and 3d. The neighbours do not largely differ from the reference images, but some of the faces are rather significantly altered in the case of RAND; for instance, the sixth column framed by dotted green box in Fig. 3c. For quantitative analysis, the Pearson's correlation coefficient between the reference image $P^{REF(k)}$ and $P^{N(k,l)}$ is evaluated given by

$$R(P^{REF(k)}, P^{N(k,l)}) = \frac{\sum_x \sum_y \left( P^{REF(k)} - \overline{P^{REF(k)}} \right)\left( P^{N(k,l)} - \overline{P^{N(k,l)}} \right)}{\sqrt{\left( \sum_x \sum_y \left( P^{REF(k)} - \overline{P^{REF(k)}} \right)^2 \right)\left( \sum_x \sum_y \left( P^{N(k,l)} - \overline{P^{N(k,l)}} \right)^2 \right)}}, \quad (3)$$

where $x$ and $y$ denotes the horizontal and vertical axis of the generated images, respectively. The similarity in proximity is characterized by the average over the neighbours and reference points given by

$$\frac{1}{KL} \sum_{k=1}^{K} \sum_{l=1}^{L} R(P^{REF(k)}, P^{N(k,l)}). \quad (4)$$

Figure 4a summarizes the values of the similarity in proximity of the images generated via the seven kinds of random sequences [(i) to (vii) in Fig. 2]. With using chaos, we observe that the similarity in proximity is enhanced, particularly with the space-domain chaotic sequences (vi), compared with the conventional uniformly distributed random numbers (i). The enhanced similarity in proximity is, however, also observed when (v) RANDN with σ = 0.2 was used, indicating that the *distribution* of the input ***Z*** is affecting the resultant property, rather than the chaotic property of the random sequences. This point will be further examined in the following discussions.

## Diversity

We quantify the diversity of the generated images by means of the multi-scale structural similarity (MS-SSIM) [14], which has been applied in the analysis of images generated by GAN in the literature [15, 16]. MS-SSIM is a multi-scale variant of perceptual similarity that has been shown to correlate well with human judgement [14]. MS-SSIM value ranges from 0 (low similarity) to 1 (high similarity). After training the generator for the seven different random sources, 10,000 images were generated by pseudorandom numbers. We calculate the MS-SSIM scores between randomly selected 1,000 pairs of images. Here we define *diversity* as [1 − *S*] where *S* is the average of all MS-SSIM values. The figure-of-merits were examined by the average of ten different trials. (See *Methods* section for details.)

With an intuitive thinking, the high proximity similarity leads to degraded global diversity while the low similarity in proximity infers higher global diversity. Indeed, as summarized in Fig. 4, the maximum and the minimum diversity are given by (i) RAND and (vi) Chaos [Space-domain] (SI: 10 ps), which coincides with the minimum and the maximum proximity similarity, respectively. However, such correspondence does *not* necessarily hold with the time-domain chaos; the similarity in proximity of time-domain chaos (ii) is the second largest, but the diversity of (ii) is also large (the third largest; almost the second largest). That is, the global diversity is not severely damaged in the case of chaos.

These observations suggest that the property inherent in the original random sequences affects the resultant generated images.

In order to further examine such properties, the sampling intervals of the original laser chaos sequences were configured from 10 to 100 ps with 10 ps interval; each of such sequence is then configured as the latent variable $Z(t)$ via the time-domain arrangement introduced above. The generated images after the training were subjected to the proximity similarity analysis. Again, ten different trials were conducted for each data set, and their average was evaluated. Here we remind that, as shown in Fig. 2(ii, d) marked by the dotted box, the time-domain correlation of $Z(t)$ exhibits a negative maximum peak value when the time lag is 50 ps. As demonstrated by the square marks in Fig. 4b, the similarity in proximity also shows a peak value at the sampling interval of 50 ps. Meanwhile, the circular marks in Fig. 4b represent the similarity in proximity when the input chaotic sequences are randomly shuffled to eliminate the temporal structure contained in the original chaos. (See *Methods* section for details of surrogate time series.) As a result, the similarity in proximity by surrogate chaos series exhibits an uncertain profile since the surrogation kills the original temporal structure; such observation supports the transformation of the time-domain property of the input random sequences to the generated images. A remark here is that these results potentially indicate that certain characteristics of random sequences used in the data generation phase, e.g., nonlinearity in chaos, can be inferred from the generated images as a kind of signature, which implies some relations to the context of security aspects of GAN [17, 18]; such points will be one of interesting future studies.

## Retrieving by chaos

Additionally, we generate the output images from the trained generator *G* by chaotic sequences (Time-domain arrangement, 50 ps interval) as well as normally distributed pseudorandom numbers with standard deviation of 0.2 (RANDN) unlike the conventional uniformly distributed pseudorandom numbers (RAND).

In Fig. 5, the random sequences used in the training are represented by different *shapes* of marks; [diamond] pseudorandom (RAND), [circle] time-domain chaos (SI: 50 ps), [triangle] pseudorandom (RANDN σ = 0.2), and [square] space-domain chaos (SI: 10 ps). From these generated models, we retrieve images by different type of random sequences which are depicted by different *colours*; [blue] RAND, [orange] Chaos, [yellow] RANDN. When the generator was trained by RAND, both the proximity similarity and the diversity are enhanced when the generator was retrieved by chaos and RANDN compared with the case of RAND. When the generators are trained by chaos and RANDN, the diversity increases both by chaos and RANDN compared with the case of RAND. At the same time, the difference between chaos- and RANDN-based retrieval is *not* evident, indicating that the distribution of the random numbers is dominant to provide such an effect. Indeed, the average of minimum Hamming distances between latent variables is about 3 in the case of RAND while those of chaos and RANDN are about 10, which would lead to superior diversity of the generated images by chaos and RANDN. Further understandings of the underlying mechanisms that distinguishes chaos and RANDN in the retrieval phase is a future study.

## Conclusion

We examine the impact of chaotic sequences for the application of artificial data generation by developing DCGAN with CelebA datasets in which experimentally observed laser chaos signals are utilized for the latent variables. The properties inherent in the chaotic signals, such as temporal correlations, are transformed to the generated data observed through the analysis of the generated data via proximity similarity analysis and diversity, which are also supported by the comparison to the cases with surrogate chaotic sequences. This study is a first step toward gaining insights into the intersections between chaos and GAN; or in a wider context, an initial exploration of novel composite systems of artificial intelligence, nonlinear dynamics, and photonic technologies.

## Methods

**Laser Chaos.** A semiconductor laser (NTT Electronics, KELD1C5GAAA) operated at a centre wavelength of 1547.785 nm is coupled with a polarization-maintaining (PM) coupler. The light is connected to a variable fibre reflector which provides delayed optical feedback to the laser, generating laser chaos. (i) Chaos 1, (ii) Chaos 2, (iii) Chaos 3, and (iv) Chaos 4 were obtained by the variable reflector by letting 210, 120, 80, and 45 μW of optical power be fed back to the laser, respectively. The length of the fibre between the laser and reflector was 4.55 m, corresponding to a feedback delay time of 43.8 ns. The output light at the other end of the PM coupler is detected by a high-speed, AC-coupled photodetector (New Focus, 1474-A) through an optical isolator and optical attenuator, sampled by a digital oscilloscope (Tektronics, DPO73304D).

**Signal processing.** The model and the parameters GAN are the ones shown in Ref. [2]. The code of GAN was built on Chainer [19]. The analysis of the generated images is performed by MATAB with Signal Processing Toolbox. We utilized the code of MS-SSIM available from [20]. Mersenne Twister is used for the pseudorandom number generator (RAND and RANDN). In the training phase (or model generation phase), 10 different random sequences were generated by different seeds, which were subjected to independent learning processes. We examine the average of the resultant figures. Regarding the cases of chaotic sequences, let **Z** be a matrix subjected to the generator where the row of **Z** is sequentially used as the latent variables. Here, the *n*-th trial of learning begins with 128× (*n* − 1) + 1-th row of **Z**. The random shuffling of the chaotic sequences is implemented by `randperm` function in MATLAB. The average of ten different random shuffling was examined for the surrogate data.

**Data availability.** The datasets generated during the current study are available from the corresponding author on reasonable request.

**Acknowledgements**

The authors thank Takashi Shinozaki for the discussion and technical supports for the study. This work was supported in part by the CREST project (JPMJCR17N2) funded by the Japan Science and Technology Agency, the Core-to-Core Program A. Advanced Research Networks and Grants-in-Aid for Scientific Research (JP17H01277 and JP16H03878) funded by the Japan Society for the Promotion of Science.

## Author Contributions

M.N. directed the project. K.K. and A.U. conducted the laser chaos experiments. M.N. and N.C. designed the analysis. M.N. and T.Y. performed the signal processing. M.N., N.C., K.K. and A.U. analysed the data. M.N. wrote the paper.

## Additional Information

Competing Interests: The authors declare no competing inter

## Additional Information

Competing interests: The authors declare no competing interests.

Correspondence and requests for materials should be addressed to M. N.

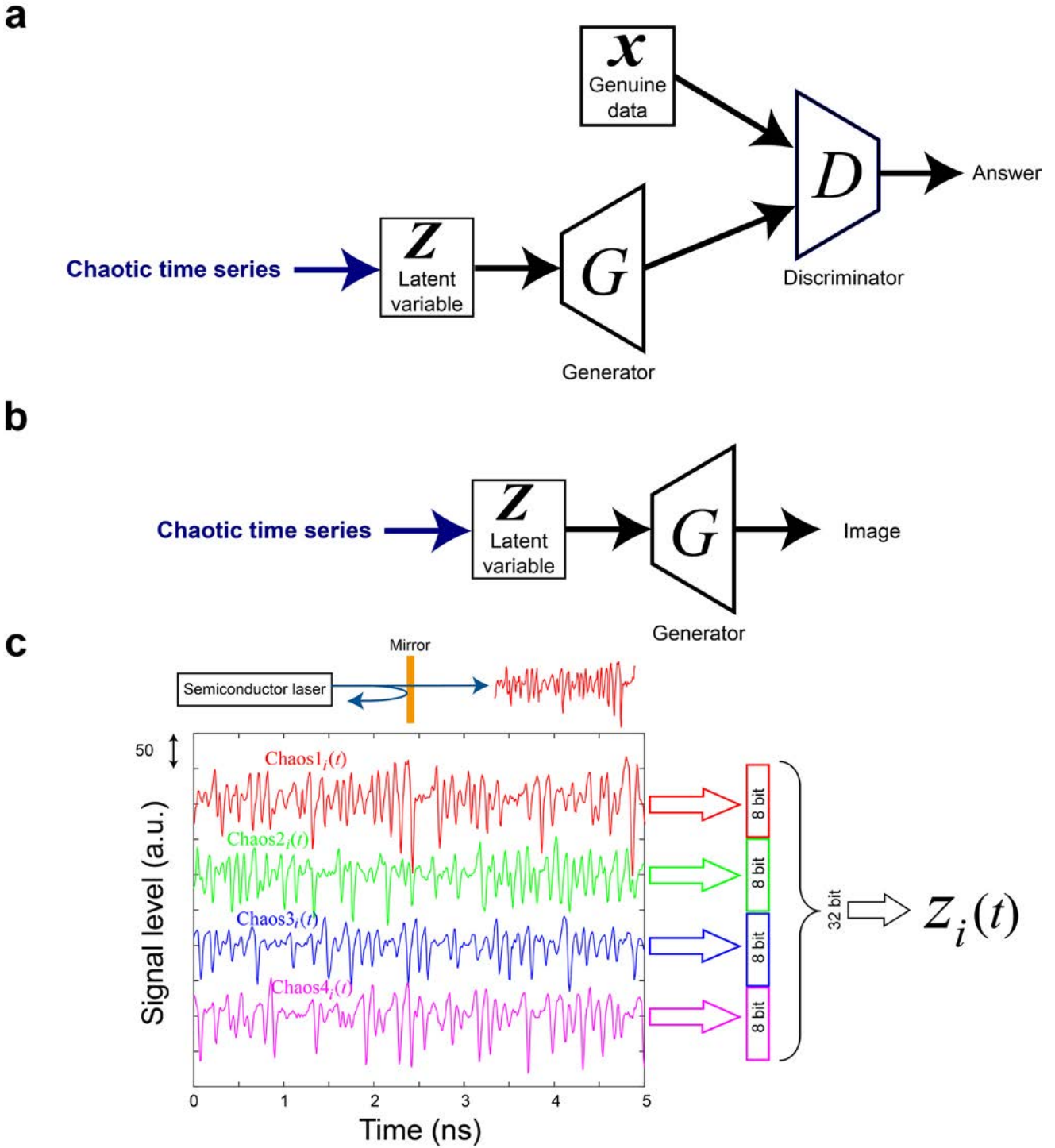

**Figure 1.** Using chaotic sequences for GAN. (**a**) Using chaotic time series in the training phase and (**b**) final data generation in GAN. (**c**) Construction of latent variables from experimentally generated laser chaos sequences.

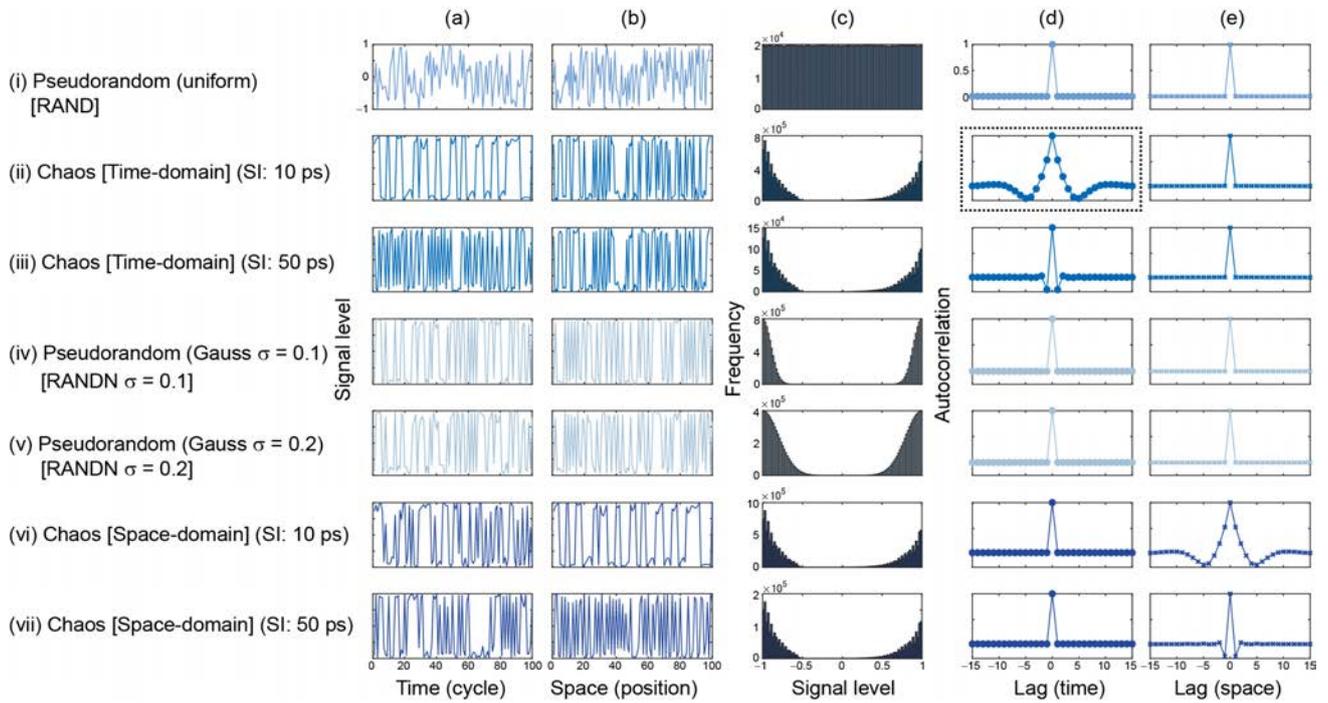

**Figure 2.** Random sequences for GAN. (i) Pseudorandom numbers (ii, iii) Chaos sequences arranged in the time domain. Sampling interval (SI) (ii) 10 ps, (iii) 10 ps (iv, v) Normally distributed pseudorandom numbers. Standard deviation (iv) 0.1, (v) 0.2 (vi) Chaos sequences arranged in the space domain. SI (vi) 10 ps, (vii) 50 ps. (a, b) Signal level profile in (a) time and (b) space domain. (c) Signal level histogram (d, e) Autocorrelation in (d) time and (e) space domain.

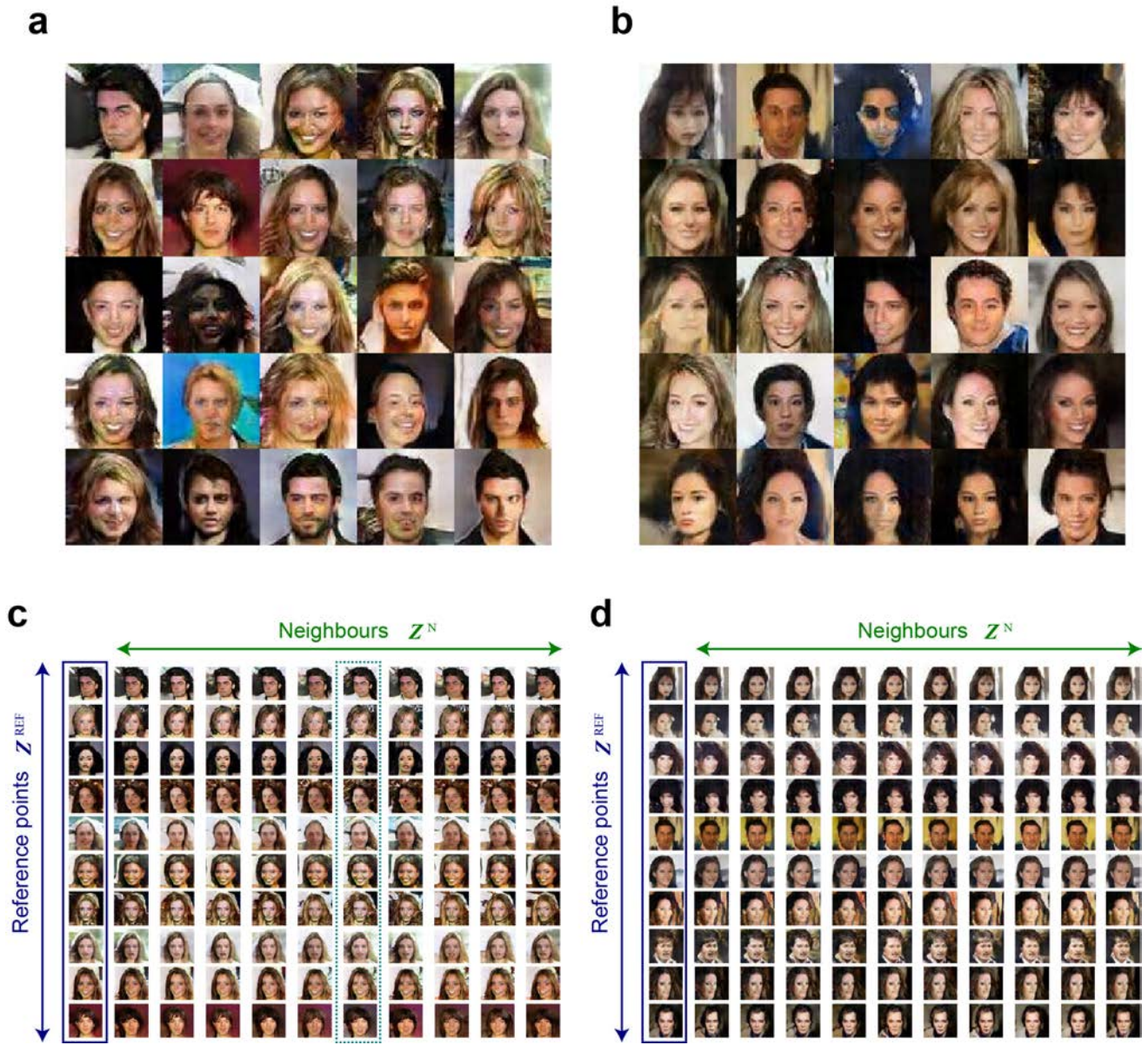

**Fig. 3.** Comparison of generated images by pseudorandom numbers and chaos. (**a, b**) Examples of generated faces images trained by (**a**) pseudorandom numbers and (**b**) chaotic sequences. (**c, d**) Proximity analysis in the generated images via training via (**c**) pseudorandom numbers and (**d**) chaos.

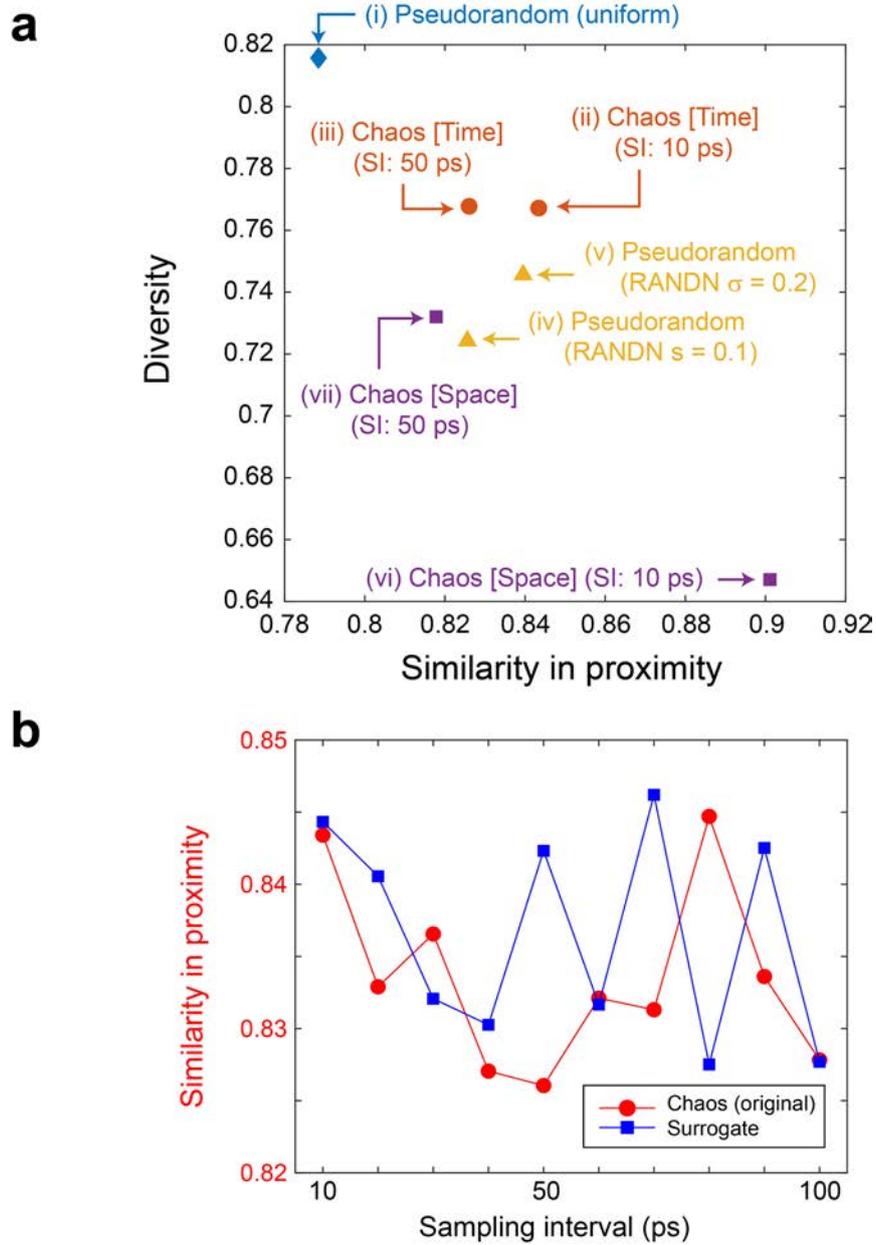

**Fig. 4.** Analysis of the generated images. (**a**) Scatter plot of the similarity of the generated images by a minute change in the latent variable, called the similarity in proximity, and the diversity of the generated images as a whole. (**b**) Similarity in proximity are characterized as a function of the sampling interval of the laser chaos sequences. It shows a minimum peak value at 50 ps, which corresponds with the negative maximum autocorrelation at 50 ps in the original laser chaos (see Fig. 2). With the surrogate chaos sequences, namely, eliminating temporal structure in the original chaos sequence, no such resulting tendency is observed.

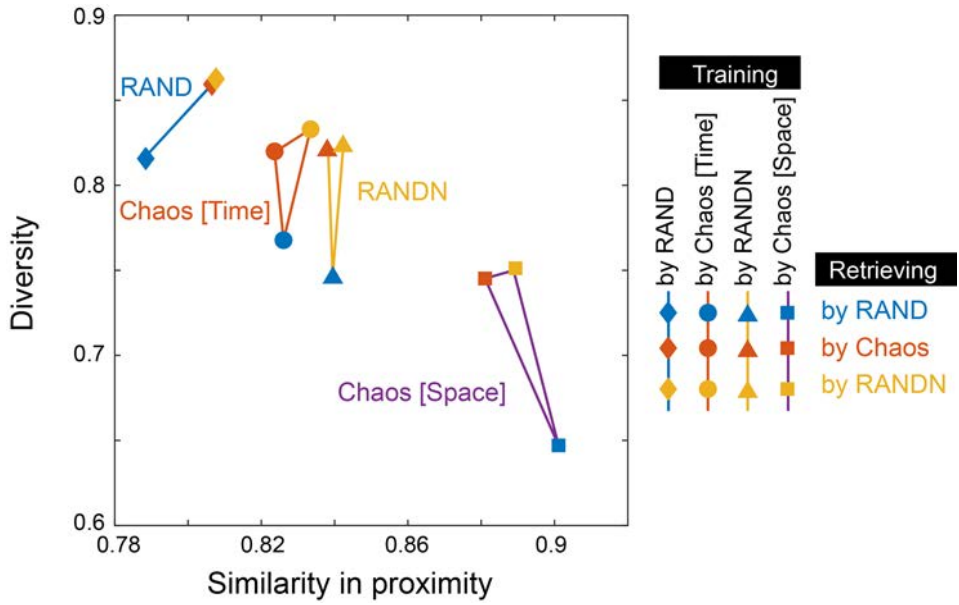

**Fig. 5.** Addressing the trained generator by chaos. Similarity in proximity and diversity when the output images are obtained by addressing the trained generator by pseudorandom numbers and chaos.